\documentclass[sigconf]{acmart}

\usepackage{graphicx}
\usepackage{amsfonts}
\usepackage{amsmath}
\usepackage{booktabs}
\usepackage{adjustbox}
\usepackage{subfig}
\AtBeginDocument{%
  \providecommand\BibTeX{{%
    \normalfont B\kern-0.5em{\scshape i\kern-0.25em b}\kern-0.8em\TeX}}}

\copyrightyear{2023}
\acmYear{2023}
\setcopyright{rightsretained}
\acmConference[SIGSPATIAL '23]{The 31st ACM International Conference on
Advances in Geographic Information Systems}{November 13--16,
2023}{Hamburg, Germany}
\acmBooktitle{The 31st ACM International Conference on Advances in
Geographic Information Systems (SIGSPATIAL '23), November 13--16, 2023,
Hamburg, Germany}\acmDOI{10.1145/3589132.3625567}
\acmISBN{979-8-4007-0168-9/23/11}

\begin{document}

\title{Learning Dynamic Graphs from All Contextual Information for Accurate Point-of-Interest Visit Forecasting}

\author{Arash Hajisafi\textsuperscript{1}, Haowen Lin\textsuperscript{1}, Sina Shaham\textsuperscript{1}, Haoji Hu\textsuperscript{2}, Maria Despoina Siampou\textsuperscript{1}, Yao-Yi Chiang\textsuperscript{2}, Cyrus Shahabi\textsuperscript{1}}
\affiliation{%
\institution{\textsuperscript{1}University of Southern California, Department of Computer Science, Los Angeles, \country{United States}}
\institution{\textsuperscript{2}University of Minnesota, Department of Computer Science and Engineering, Twin Cities, \country{United States}}
}
\email{{hajisafi, haowenli, sshaham, siampou, shahabi}@usc.edu, {huxxx899, yaoyi}@umn.edu}

\renewcommand{\shortauthors}{Hajisafi, et al.}

\begin{abstract}
Forecasting the number of visits to Points-of-Interest (POI) in an urban area is critical for planning and decision making in various application domains, from urban planning and transportation management to public health and social studies. Although this forecasting problem can be formulated as a multivariate time-series forecasting task, current approaches cannot fully exploit the ever-changing multi-context correlations among POIs. Therefore, we propose Busyness Graph Neural Network (BysGNN), a temporal graph neural network designed to learn and uncover the underlying multi-context correlations between POIs for accurate visit forecasting. Unlike other approaches where only time-series data is used to learn a dynamic graph, BysGNN utilizes all contextual information and time-series data to learn an accurate dynamic graph representation. By incorporating all contextual, temporal, and spatial signals, we observe a significant improvement in our forecasting accuracy over state-of-the-art forecasting models in our experiments with real-world datasets across the United States.
\end{abstract}






\maketitle

\section{Introduction}
Point-of-Interest (POI) data is a treasure trove of information, providing geographical locations, entity names, and types of places of interest, such as the Eiffel Tower (landmark) or a Starbucks coffee shop. Predicting the number of visitors to a POI at a specific time offers valuable insights into collective social and mobility behavior~\cite{hao2023mbp}. It has numerous practical applications, such as traffic flow analysis, epidemic spread prediction, and travel demand estimation (e.g., for ride-hailing apps). Take the prediction of epidemic spread as an example: during the COVID-19 pandemic, accurate predictions of the number of visits to grocery stores in a particular neighborhood could inform policies on store closures or visitor restrictions to control the spread of the virus (e.g.,~\cite{chang2021mobility}).

Forecasting POI visiting patterns is a complex task due to the ever-changing nature of human mobility behavior. External factors such as rush hours, seasonal traffic fluctuations, weather, holidays, and planned or unexpected events, transient, such as a football game or long-term, such as the COVID-19 pandemic, all contribute to this unpredictable behavior. One way to improve forecasting accuracy is to exploit the similarities between POIs, a non-trivial task. It involves identifying and effectively combining correlations from different signals such as past visiting patterns, geographical locations, semantic similarities (e.g., shared POI attributes such as cuisine types of restaurants), and taxonomic distances (e.g., similar visitation trends between categories such as ``restaurant'' and ``bar''). These signals also tend to change over time, making it challenging to infer similarity between two POIs even when considering correlations based on past POI visit numbers.

While predicting future POI visit numbers can be formulated as a time-series forecasting task, there are several limitations to existing methods. Classical time-series forecasting methods like ARIMA~\cite{makridakis1997arma} and VAR \cite{lutkepohl2005new} rely on assumptions of linearity and stationarity, which do not hold in complex real-world scenarios and fail to capture long-term dependencies. Recently, deep learning has achieved impressive results in various tasks, such as image classification \cite{lin2021semifed,lin2021integer}, natural language processing \cite{vaswani2017attention}, and ensuring privacy and fairness \cite{shaham2023holistic,shaham2022models}. However, typical deep learning models such as RNN-based approaches (e.g., LSTM~\cite{hochreiter1997long} and GRU~\cite{cho2014learning}), while capturing {\em intra} time-series dependencies, both short-term and long-term, cannot exploit relationships across time series. One way to capture these {\em inter} time series correlations is to conceptualize the problem as a graph, where nodes are POIs and edges (and edge weights) capture their interdependencies, and then apply Graph Neural Networks (GNNs) on the resulting graph. The challenge here is how to build a representative graph.  

Toward this end, the GNN approaches can be divided into {\em Static} and {\em Dynamic} categories. The static GNNs build the graph once based on input feature vectors using predefined similarity measures, often derived from specific domain knowledge~\cite{li2017diffusion, yu2017spatio, seo2018structured}. These models emphasize less on graph construction, but focus on graph convolution techniques and processing methods. For example, DCRNN~\cite{li2017diffusion} builds a simple static graph of traffic sensors based on their road-network distances and then passes it to Graph Diffusion Convolution with a sequence-to-sequence architecture. The second category learns a dynamic graph representing the time-varying relationships between variables~\cite{velivckovic2017graph, wang2022hagen, cao2020spectral, zheng2020gman}. For instance, StemGNN models latent correlations between time-series windows to generate a time-varying graph on which it applies Spectral Graph Convolution~\cite{kipf2016semi}, which uses Graph Fourier and Discrete Fourier transforms to capture time series correlations.

For POI visit forecasting, visit patterns of a single POI change over time for various reasons, which makes the dynamic GNNs a suitable solution. However, in certain time windows, say during holidays or COVID, the visit patterns of two semantically (and/or geographically) far POIs may look similar. 
 Conversely, due to some temporary events, e.g., remodeling, two similar POIs may have different visit patterns. Therefore, our goal is to build a comprehensive dynamic graph using all contextual information robust to time variances (dynamic) or predefined node similarities~(static).

Consequently, our proposed Busyness Graph Neural Network (BysGNN) builds a dynamic graph by capturing POIs' spatial correlations, intra-series dependencies in individual visit patterns, inter-series dependencies across visit patterns, semantic similarity, and taxonomic proximity. This is achieved through a robust gated attention mechanism and an effective thresholding strategy. The gated attention mechanism integrates semantic-based and distance-based similarity matrices as a gate to determine the extent to which similarity scores between time series should be considered to generate the graph's adjacency matrix. Subsequently, the thresholding mechanism eliminates noisy relationships to improve the graph representation's accuracy. Another challenge is that, while semantic similarities are not dynamic, they are task-dependent and are unknown a priori. To address this, BysGNN uses a pre-trained language model to obtain initial semantic embedding based on the textual description of POIs and fine-tunes them based on the forecasting task to account for accurate semantic similarities as the network trains.

We conducted extensive experiments with real-world datasets and compared BysGNN with some naive baselines and state-of-the-art static and dynamic GNNs. The results demonstrate the superiority of BysGNN in effectively building dynamic graphs incorporating information from various contextual and past visit signals. The experiments show that previous dynamic GNNs that rely solely on visit pattern similarity require many similar POIs to perform well. In contrast, static GNNs perform better with fewer POIs. Interestingly, a naive baseline outperforms static GNNs but falls short compared to dynamic GNNs and BysGNN, underscoring the significance of a dynamic graph structure. Our ablation study validates the positive impact of simultaneously considering the inter-time-series, semantics, spatial, and taxonomic similarities on forecasting accuracy, with semantics showing the most substantial improvement. Finally, we illustrate example cases where other dynamic GNNs consider two POIs to be similar (or different) due to visit pattern similarities (or dissimilarities) in a recent time window, resulting in an inaccurate adjacency matrix. In contrast, BysGNN’s gated adjacency matrix shows a high influence from long-term semantic/geographical similarities (or dissimilarities), resulting in accurate forecasting.

The rest of the paper is organized as follows. We review the related work for time-series forecasting in Section \ref{related-work}. Section \ref{problem-statement} formally defines the problem of forecasting POI visit numbers. We describe our BysGNN framework in Section \ref{methodology}. Finally, we present our experimental setup, datasets and results in Section \ref{experiments} and conclude the paper in Section \ref{conclusion}.

\section{Related Work}
\label{related-work}
Time-series modeling has long been a prominent area of academic research, leading to the development of a wide variety of forecasting methods. These methods can be broadly categorized into univariate and multivariate time series techniques. Univariate technologies focus on analyzing single observations recorded sequentially without considering correlations between different time series variables~\cite{rather2015recurrent,liu2016online,salinas2020deepar}. For instance, the ARIMA family of methods assumes a linear relationship, where predictions are weighted linear sums of past observed values. Salinas et al. propose a forecasting method based on autoregressive recurrent neural networks, which models the probability distributions of the variable in the future \cite{salinas2020deepar}. In contrast, multivariate time series techniques aim to capture interactions and co-movements among a group of variables \cite{wu2020connecting,audibert2020usad,nguyen2023evaluation}. For example, Zerveas et al. present a novel framework based on a transformer encoder that extracts dense vector representations of multivariate time series \cite{zerveas2021transformer}.

Graph Neural Networks (GNNs)~\cite{gori2005new,scarselli2008graph} have emerged as powerful machine learning models for modeling non-Euclidean data represented by graphs. In recent years, the application of GNNs in multivariate time series forecasting has witnessed significant success across various domains.
One notable application is DCRNN~\cite{li2017diffusion}, which models traffic flow as a diffusion process on a directed graph, effectively capturing spatial dependence between sensors for traffic forecasting tasks. However, such approaches often rely on predefined correlations between components to pre-construct a graph, which remains fixed during training and testing. Another approach, HAGEN \cite{wang2022hagen}, introduces a graph convolutional recurrent network that dynamically captures crime correlations between regions and temporal crime dynamics for crime forecasting. Additionally, StemGNN \cite{cao2020spectral} successfully captures inter-series correlations and temporal dependencies jointly for multivariate time series forecasting.

Dynamic GNNs have been used extensively for epidemic forecasting, e.g., COLA-GNN \cite{deng2020cola} proposes a novel GNN-based framework with a location-aware attention mechanism to capture spatiotemporal dependencies, enabling accurate long-term predictions. STAN~\cite{gao2021stan} is another prediction framework that utilizes graph attention networks to incorporate interactions between similar locations, enhancing its accuracy in the prediction of pandemics. Moreover, CausalGNN~\cite{wang2022causalgnn} adopts an attention-based approach to learn a combined spatiotemporal and causal latent embedding from disease dynamics and epidemiological context, facilitating precise forecasting of daily new COVID-19 cases.

In contrast to most GNN-based frameworks that rely on recurrent neural networks to capture temporal dependencies, Choi et al. introduce a novel approach \cite{choi2022graph} by integrating neural-controlled differential equations with graph convolution processing technology for spatiotemporal forecasting.

Despite the abundance of graph-based modeling approaches in time series forecasting, these approaches cannot effectively fuse multiple contextual sources, which is the focus of BysGNN, and thus can be considered an orthogonal approach to these models.

\section{Problem Formulation}\label{problem-statement}
This section provides the preliminaries and a formal definition of the problem of forecasting POI visit numbers.

\begin{definition}
(\emph{Multi-Context Correlations}). Multi-context correlations refer to the latent relationships among POIs influenced by various contextual factors, such as time of day, day of the week, distance, and events. In POI visiting number forecasting, these correlations include spatial (closeness of geographic locations), temporal (the changes in visit patterns over time and the dependencies between visit patterns of different POIs), semantic (similarity of POI attributes, such as POI types), and taxonomic (the general semantic categories of POIs, representing the high-level visiting trend) correlations. Leveraging these multi-context correlations can enhance the accuracy of forecasting frameworks for POI visits.
\end{definition}


\begin{definition}
(\emph{Busyness Graph}).
    We define a \emph{Busyness Graph} network $G=(V, A)$ where $V$ is a set of $|V| = N$ nodes, and each node corresponds to a specific POI (e.g., a restaurant) or a category of POIs (e.g., all restaurants). We denote the $A\in \mathbb{R}^{N\times N}$ as the adjacency matrix, where
$a_{ij} > 0$ indicates that there exists an edge connecting nodes $v_i$ to $v_j$, and $a_{ij}$ indicates the strength of this edge which shows the amount of influence that $v_i$ has on the forecasts of $v_j$. The adjacency matrix $A$ is dynamically updated based on the multi-context correlations and captures the most recent knowledge about the interaction between POI nodes.

\end{definition}

    

\begin{definition}
(\emph{POI Visit Forecasting Problem}). Given $X=(x_1,...,x_N) \\ \in \mathbb{R}^{N\times T}$ as the input time series that represents the hourly visit numbers to each POI for a time window of $T$ steps, and $U=(u_1,..,u_N)\in\mathbb{R}^{N\times J}$ as the set of $J$ attributes (e.g., category and name) of each POI, our goal is to generate and utilize the dynamic Busyness Graph $G$ to find $Y=(y_1,...,y_N)\in \mathbb{R}^{N\times H}$, which shows the future visit numbers for the next $H$ time steps for each POI.
\end{definition}

\begin{figure*}[t]
  \centering
          \begin{adjustbox}{width=\linewidth, center}
\includegraphics{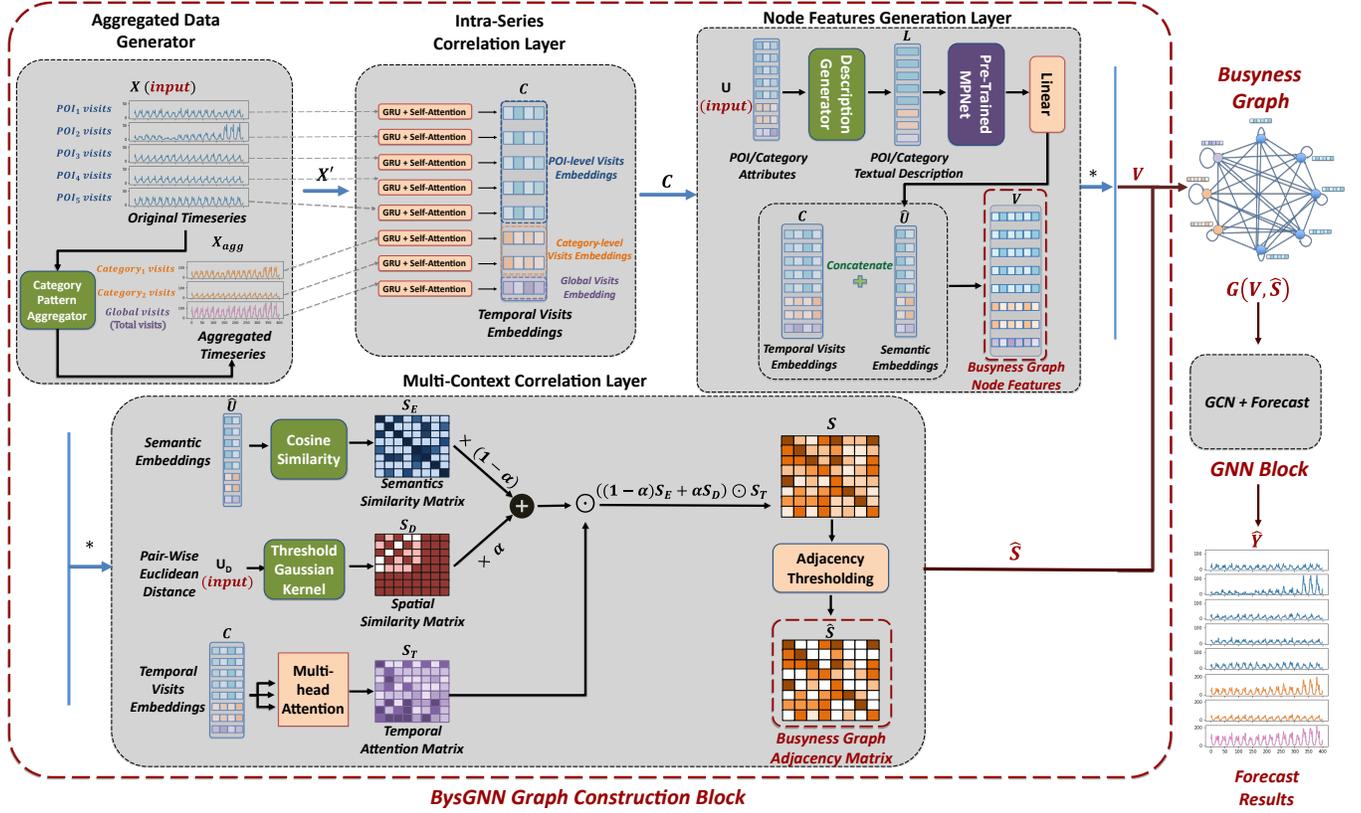}
        \end{adjustbox}
  \caption{\textbf{BysGNN Overall Framework}}
  \vspace{-10pt}
  \label{frame-arch}
\end{figure*}

\section{Busyness Graph Neural Network}\label{methodology}
\subsection{Overview}
This section describes the BysGNN (Busyness Graph Neural Network) framework to tackle the problem of forecasting POI visiting numbers. For each forecasting horizon, BysGNN first generates a dynamic graph by exploiting all contextual information. The graph edges and their strengths represent the multi-context correlations between POIs. Then BysGNN uses the dynamic graph for accurate forecasting.

The overall architecture of BysGNN is presented in Figure \ref{frame-arch}, which consists of two main blocks. The first block, referred to as the \textbf{\emph{BysGNN Graph Construction Block}} (sec \ref{pre-graph-block}), is responsible for building the dynamic \textbf{\emph{Busyness Graph}}, which captures the multi-context correlations among POIs. The second block called the \textbf{\emph{GNN Block}} (sec \ref{sec:gnn-block}), performs the convolution operation on the Busyness Graph and generates node embeddings used to make forecasts.

The BysGNN's \emph{Graph Construction Block} starts by feeding the input time series to the \emph{\textbf{Aggregated Data Generator}} module (sec \ref{agg-data-gen}), which generates new aggregated time-series data based on a predefined measure (POI taxonomy in our case) and adds them to the original input. This step allows the model to learn the \emph{taxonomic correlations} in the following steps. 
The augmented time-series data are passed through the \emph{\textbf{Intra-Series Correlation Layer}} (sec \ref{sec:intra-corr-layer}), which learns the temporal dependencies in individual time series (i.e., intra-series correlations) and generates temporal embeddings summarizing the time-series information for the given time window.
The temporal embeddings are then passed to the \emph{\textbf{Node Features Generation Layer}} (sec \ref{node-feat-gen}), which assigns a node to each individual time series and generates feature vectors for each node in the graph. We call the nodes corresponding to aggregated time series ``meta-nodes'' and the nodes corresponding to individual POIs ``POI nodes.'' This layer first builds semantic embeddings based on POI attributes (such as categories and names) that allow the model to learn \emph{semantic similarities} between nodes. The next step concatenates the semantic embeddings with the previous temporal embeddings to form the node feature vectors.

The next step involves passing the semantic and temporal embeddings into the \emph{\textbf{Multi-Context Correlation Layer}} (sec \ref{sec:multi-context-layer}). This layer is crucial as it generates the adjacency matrix of BysGNN's dynamic graph, which reflects the dependencies among POI nodes and meta-nodes across multiple contexts, including inter-series relationships across time series, the spatial proximity of POIs, and semantic similarities between POI and meta nodes.
To ensure reliable inter-series correlation scores, BysGNN incorporates a gating mechanism that combines pairwise spatial and semantic similarity scores as a gate to allow the flow of inter-series correlation scores. This mechanism enables the model to effectively use the inter-series similarity scores for forecasting by considering the spatial and semantic context between POIs.

This approach results in a more robust and sparser adjacency matrix, preventing over-smoothing by retaining strong relationships while reducing the impact of weak and noisy relationships. The layer also utilizes a \emph{case amplification} technique to threshold the adjacency matrix and remove noise. 


After generating the node features and adjacency matrix in the previous layers, BysGNN creates the dynamic \emph{Busyness Graph}. Busyness Graph is then passed to the \emph{GNN Block} for further processing, where a Graph Convolution layer is applied to obtain forecasts based on the final node embeddings.



\subsection{BysGNN Graph Construction Block}
\label{pre-graph-block}
\subsubsection{\textbf{Aggregated Data Generator}}
\label{agg-data-gen}

This layer generates and adds new aggregated time series based on POI types (categories) to the original time series data, allowing the model to learn \emph{taxonomic correlations}. For example, POIs in the Gas Station category might follow a similar visit pattern. At the same time, the visit pattern of POIs in the Restaurant category could also be similar to the aggregated visiting patterns of POIs in the Gas Station category. On top of this, the degree to which the POIs in the same category follow the same pattern as the aggregated pattern of that category differs vastly between different POIs. As a result, the taxonomic correlation that BysGNN defines consists of correlations between patterns at the same aggregation level and correlations between visiting patterns from different levels of aggregation. In the POI visit forecasting problem, BysGNN considers three different aggregation levels corresponding to individual POIs' visit patterns, POI categories' visit patterns, and the Global visits pattern (an additional time series generated by aggregating the visits to all POIs), respectively. Adding these aggregated time series allows BysGNN to learn such taxonomic correlations in the next step.


Specifically, given $X \in \mathbb{R}^{N\times T}$ as the original time-series input (with $N$ as the number of time series and $T$ as the window size for each series), and $K$ as the set of all input POI category types, this module generates $|K|$ new time-series data by aggregating individual POI time-series based on their categories, and an additional time series by aggregating all the time series data together (\emph{Global} visits time series) and adds them to the original input. As a result, the module output will be $X' \in \mathbb{R}^{(N+|K|+1)\times T}$. The aggregation is defined as a function $f_{agg}$ such that:
$f_{agg}:\mathbb{R}^{N'\times T'} \rightarrow \mathbb{R}^{T'}$ (summation, in our case).


\subsubsection{\textbf{Intra-Series Correlation Layer}}
\label{sec:intra-corr-layer}
\begin{figure}[t!]
  \centering
  \includegraphics[width=\linewidth]{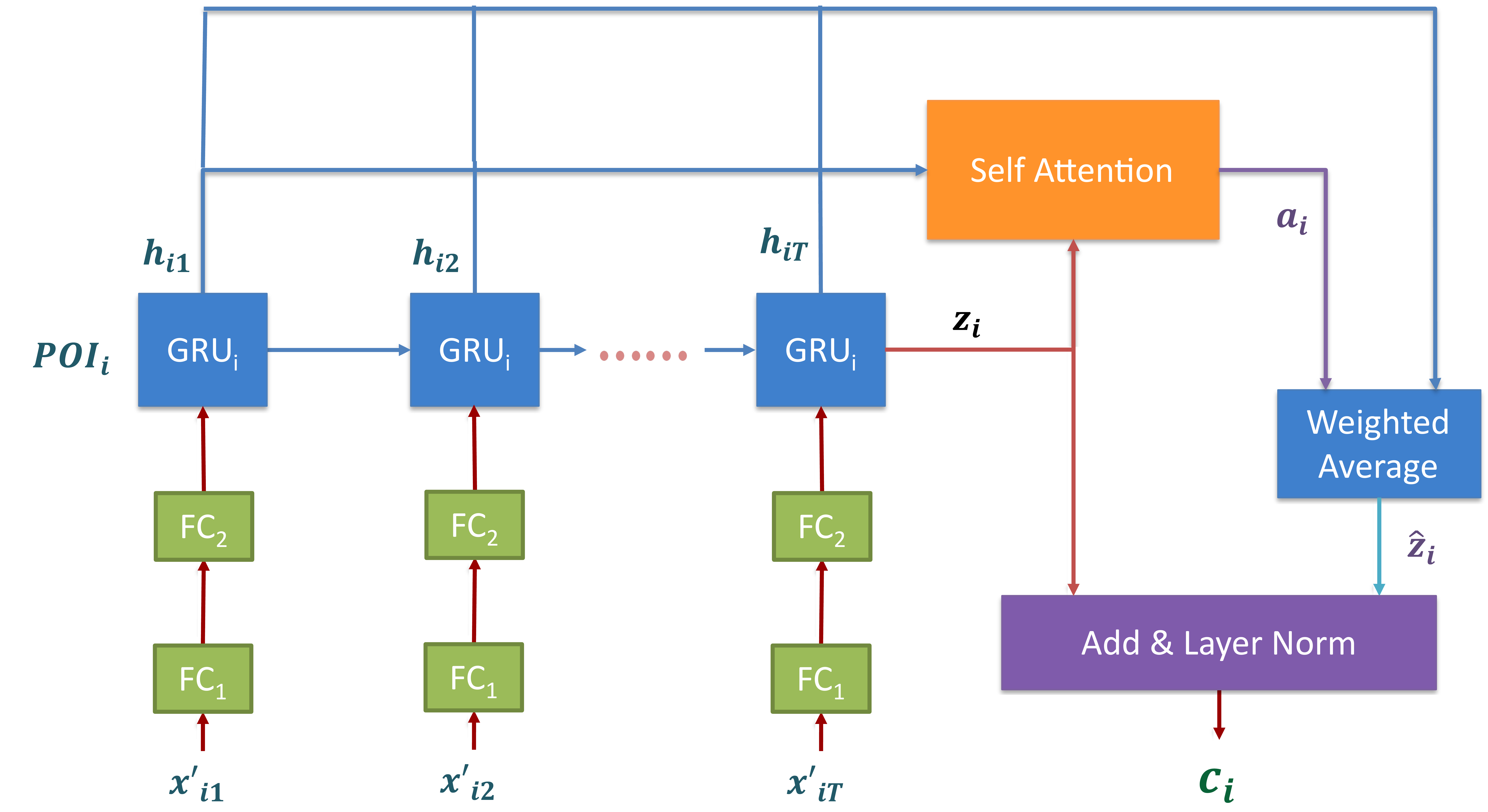}
  \caption{Intra-Series Correlation Layer}
  \label{intra-corr}
\end{figure}

This layer is responsible for learning the intra-series dependencies in individual time series and generating embedding vectors summarizing the time-series data for the given window.
The layer receives the augmented time series data $X' \in \mathbb{R}^{(N+|K|+1)\times T}$ as input and generates time-series embeddings $C\in \mathbb{R}^{(N+|K|+1)\times M}$, where $M$ represents the embedding dimension. The layer consists of separate $N+|K|+1$ GRU weight matrices to learn the temporal intra-series dependencies for each time-series data independently. Figure \ref{intra-corr} illustrates the process of generating these embeddings for each series. It is important to note that even though separate GRU units are used for different time series,
the size of these GRU units is relatively small, and the number of parameters and training time will not be significantly different compared to using one big GRU unit.

Each time series $x'_i=(x'_{i1},...,x'_{iT})\in \mathbb{R}^{T}$ holds the visiting numbers for a sequence of $T$ time steps. First, BysGNN maps each time-series sequence to the space of higher dimensions by passing data points in the series through two linear layers and getting $\hat{X_i}=(\hat{x}_{i1},...,\hat{x}_{iT}) \in \mathbb{R}^{T\times D}$. Then, it feeds the resulting sequence $(\hat{x}_{i1},...,\hat{x}_{iT})$ to the $i$-th GRU unit and gets the GRU states $H_i=(h_{i1},...,h_{iT}) \in \mathbb{R}^{T\times M}$, with $z_i=h_{iT}$ being the output of the GRU unit.

Inspired by \cite{qin2017dual}, instead of using only the GRUs' output $z_i$ as the summary of the window sequence for $i$-th time series in $X'$, which is prone to be affected by the final observations in the window more than it should, BysGNN utilizes a self-attention mechanism to assign weights to GRU states at different timestamps based on the importance of that state to pay more attention to the more important timestamps. 
Therefore, BysGNN passes all hidden states $(h_{i1},...,h_{iT})$ and the final GRU state $z_i$ to a self-attention component. This component calculates the attention scores $a_{it}$ for each hidden state using the following equation: 
\begin{equation}
    a_{it} =
    softmax(tanh(W_a[h_{it}\mathbin\Vert z_i])
\end{equation}
For calculating the attention score vector $a_i$, containing the attention score values $(a_{i1},...,a_{iT})$ corresponding to each hidden state, BysGNN first concatenates each vector $h_{it}$, which contains the hidden states for timestamp $t$, with the final GRU state $z_i$. Then, the concatenated vector passes through a linear layer with $W_a$ as the weights matrix and a $tanh$ activation is applied. Finally, to ensure that all attention scores remain between 0 and 1 with the attention score vector summing to 1, BysGNN passes the intermediate results through a $softmax$ layer.

BysGNN then calculates the weighted average $\hat{z}_i$ of hidden states using the attention scores vector $a_i$. Finally, it adds $z_i$ with $\hat{z}_i$ and passes the result to a layer normalization unit \cite{ba2016layer} to obtain the final temporal embedding $c_i \in \mathbb{R}^{M}$ for $i$-th time series. BysGNN puts together the embeddings for all time-series data to obtain $C \in \mathbb{R}^{(N+|K|+1)\times M}$.

\subsubsection{\textbf{Node Features Generation Layer}}
\label{node-feat-gen}

This layer is responsible for building the node features for BysGNN's dynamically generated graph by combining embeddings for time series and semantics of POIs.

To achieve this, given $u_i = (u_{i1},...,u_{iJ})\in \mathbb{R}^{J}$ as the vector of $J$ attributes for $i$-th time series, BysGNN first generates a sentence describing the corresponding POI (or POI category if the time series is an aggregated one). The attributes used include POI names, addresses, working hours, phone numbers, and top and sub-categories. Sentence generation involves predefined templates for POIs and POI categories (for meta-nodes), which are filled automatically with the corresponding attributes. For example, the following sentence describes a POI node (with italic text indicating the specific attributes of that node): 

``This point of interest is \emph{Simon mall} located at \emph{5085 Westheimer Rd, Houston, TX, 77056}. It is open for business during \emph{Monday – Friday: 10:00 - 19:00, Saturday 10:00 – 17:00, and closed on Sunday}. It can be contacted by phone at \emph{(213)538-XXXX}. The location belongs to the top-category \emph{Lessors of Real Estate}, with the sub-category \emph{Malls}.'' Similarly, the following description is generated for a meta-node: ``This is the meta-node representing all the points of interest in \emph{Houston} that belong to the top category \emph{Spectator Sports}.'' Additionally, it is worth noting that the generation of sentence descriptions is not limited to our specific POI dataset. If necessary, sentence templates can be easily customized and populated with attributes from alternative POI datasets.

Next, an intermediate embedding $u'_i$ is obtained by tokenizing and passing the generated sentence through a pre-trained MPNet language model \cite{song2020mpnet}. Although this language model is optimized to achieve state-of-the-art performance in semantic similarity tasks, it is not specifically trained to represent POI semantics accurately in the POI visit forecasting task. To address this limitation, BysGNN fine-tunes the intermediate embedding using a linear layer during training. The resulting final semantic embedding $\hat{u_i}\in \mathbb{R}^{P}$ is tailored to capture time-series semantics and has a dimensionality of $P$.

The final feature vector $v_i$ for the $i$-th time series is obtained by concatenating the temporal embedding $c_i\in \mathbb{R}^{M}$ from the output of the Intra-Series Correlation Layer with the semantic embeddings $\hat{u}_i \in \mathbb{R} ^ P$: \hspace{3pt} $v_i = c_i \mathbin\Vert \hat{u}_i \in \mathbb{R} ^ {M+P}$.


The Node Features Generation Layer block in Figure \ref{frame-arch} shows this process. By combining individual node feature vectors, the node feature matrix $V=(v_1,...,v_{N+|K|+1})\in \mathbb{R}^{(N+|K|+1)\times (M+P)}$ is obtained.

\subsubsection{\textbf{Multi-Context Correlation Layer}}
\label{sec:multi-context-layer}

The Multi-Context Correlations layer plays a crucial role in BysGNN's dynamic graph generation process. This layer is responsible for generating the adjacency matrix for the dynamic graph structure by capturing the multi-context dependencies among POI nodes and meta-nodes.

To achieve this, the layer takes in the semantic embeddings $\hat{U}\in \mathbb{R}^{(N+|K|+1)\times P}$ and the time-series embeddings $C\in \mathbb{R}^{(N+|K|+1)\times M}$ from the previous layer, as well as the pairwise Euclidean distance $U_D\in \mathbb{R}^{N\times N}$ between POI nodes from the input. It then generates three similarity matrices: semantic similarity matrix $S_E$, spatial similarity matrix $S_D$, and attention matrix $S_T$, each with a dimensionality of $\mathbb{R}^{(N+|K|+1)\times (N+|K|+1)}$.


To create the semantic similarity matrix $S_E$, BysGNN calculates the pairwise cosine similarity scores between the semantic embedding vectors in $\hat{U}$.
Cosine similarity is used as the similarity metric because it is normalized and effectively captures the degree of alignment between the semantic meanings of embedding vectors.

To generate the spatial similarity matrix $S_D$, BysGNN first passes the pairwise Euclidean distance matrix $U_D$ through a thresholded Gaussian kernel \cite{shuman2013emerging} to obtain $U'_D \in \mathbb{R}^{N\times N}$:

\begin{equation}
U'_{D}(i, j) = \begin{cases}
exp(-\frac{U_{D}^{2}(i, j)}{\sigma ^ 2}), &\text{if } U_{D}(i, j) < \tau \cr
0, &\text{otherwise}\
\end{cases}
\end{equation}

where $U_{D}(i, j)$ is the Euclidean distance between the $i$-th and $j$-th POI nodes, $\sigma$ is the standard deviation of the distances in $U_D$, and $\tau$ is a predefined threshold for sparsity. This process provides distance-similarity scores between POI nodes. As meta-nodes are not physical locations with geo-coordinates, BysGNN considers the distance similarity score within meta-nodes and between meta-nodes and POI nodes to be $1$. This ensures that the lack of geo-coordinates for meta-nodes does not impact the learned relationship between them and POI nodes. BysGNN builds the distance similarity matrix $S_D\in \mathbb{R}^{(N+|K|+1)\times (N+|K|+1)}$ such that the first $N$ elements for the first $N$ columns are the same as the corresponding elements in $U'_D$ and the rest are set to $1$.

To create a similarity matrix for time-series windows of different nodes, BysGNN utilizes Multi-head Attention, as outlined in~\cite{vaswani2017attention}, which generates an attention matrix $S_T$ representing pairwise correlation scores between time-series embeddings. The temporal embedding vectors are passed to the Multi-Head Attention unit and $S_T$ is constructed using the following operations:

\begin{equation} \label{eq:mhead-att}
S_T=MultiHead(Q,K,V)=(head_1\mathbin\Vert…\mathbin\Vert head_l)W^O 
\end{equation} 
\begin{equation} head_i=Attention(QW_i^Q,KW_i^K,VW_i^V) 
\end{equation} 
\begin{equation} Attention\left(Q,\ K,\ V\right)=Softmax\left(\frac{QK^T}{\sqrt{M}}\right)V 
\end{equation}

Here, the matrix $C$ is used as the matrix of Keys $K$, Queries $Q$, and Values $V$ simultaneously. $l$ represents the number of attention heads, while $W^O$, $W_i^Q$, $W_i^K$, and $W_i^V$ are learnable weight matrices. Moreover, $M$ refers to the dimension of each temporal embedding $c_i$. Equation \ref{eq:mhead-att} yields the attention scores matrix $S_T$.

BysGNN then utilizes a gate that combines the weighted average of semantic similarities ($S_E$) and spatial similarities ($S_D$) to control the flow of information in time series similarities ($S_T$) and create the un-thresholded adjacency matrix ($S$). Specifically, the gate is formulated as follows: 
\begin{equation} 
S_{Gate} = (1-\alpha)S_E + \alpha S_D 
\end{equation} 

Where $\alpha$ is a learnable parameter between 0 and 1 that adjusts the balance between the impact of spatial and semantic similarities based on the specific POI visits forecasting task. The Hadamard product operator ($\odot$) is then applied to the gate and the time series similarities matrix, resulting in the un-thresholded adjacency matrix ($S$) as follows:
\begin{equation} 
S = S_{Gate} \odot S_T 
\end{equation}


BysGNN's gating mechanism helps to preserve strong long-term relationships and penalize noisy relationships between distant or semantically dissimilar nodes.

Finally, to filter out previous noisy relationships, BysGNN applies a thresholding step to the adjacency matrix $S$. This is achieved by first normalizing each row of the attention matrix and then transforming the values using a case amplification power function to make it easier to differentiate between small and large values. This significantly reduces the impact of small values over those of larger ones. Next, a predefined threshold value $\eta$ is applied to the amplified matrix to obtain a binary mask.
The resulting binary mask is then applied to the un-thresholded adjacency matrix $S$ to obtain the final adjacency matrix $\hat{S}$ as follows:
\begin{equation}
\hat{S}_{ij} = \begin{cases}
S_{ij}, & \text{if } (\frac{S_{ij}}{\max(S_{i})})^p > \eta \cr
0, & \text{otherwise}
\end{cases}
\end{equation}

where $\max(S_{i})$ is the maximum value in the $i$-th row of $S$, $p$ is the exponent of the case amplification function and $\eta$ is the predefined threshold value. BysGNN utilizes $\hat{S} \in \mathbb{R}^{(N+|K|+1)\times (N+|K|+1)}$ as the adjacency matrix for the generated Busyness Graph. This process is illustrated in the Multi-Context Correlation Layer block of Figure~\ref{frame-arch}.

\subsection{GNN Block}
\label{sec:gnn-block}
After completing every step in the BysGNN Graph Construction Block (as described in section \ref{pre-graph-block}), BysGNN constructs the Busyness Graph $G(V,\hat{S})$ for the given time window by combining the node features ($V$) and the derived adjacency matrix ($\hat{S}$). This dynamically generated graph has already captured the underlying multi-context correlations between POIs. 

Next, BysGNN incorporates three Graph Convolutional Network (GCN) layers from \cite{kipf2016semi} for message passing within the Busyness Graph. Using the selected convolutional layers ($F_{GNN}$), the final node representations $V'$ are generated as follows:
\begin{equation}
V' = F_{GNN}(G(V,\hat{S})) \in \mathbb{R}^{(N+|K|+1)\times M^*}
\end{equation}
Here, $M^*$ represents the node embedding's dimension. BysGNN then passes the concatenation of $V'$ and the original node features $V$ through a fully connected linear layer to produce the final forecasts.

\section{Experiments and Discussions}\label{experiments}
This section describes our experimental setup and methodology. Details related to hardware and software setup, evaluation metrics, and hyper-parameters are available in the appendix \ref{app:exp-details}.

\subsection{Data Description}
To evaluate the accuracy of our model in forecasting hourly visitor numbers, we utilized the POI data and hourly visitation patterns datasets provided by SafeGraph \cite{safegraph}, a commercial data provider that compiles its datasets using phone GPS locations and open government data. Our experiments were conducted in five different cities and involved two data regimes (large and small) that spanned the period from January 1, 2019, to February 4, 2020. In the large data regime, we considered hourly visit counts for the top 400 most visited POIs in each city, while the small data regime focused on the top 40 most visited POIs. This allowed us to examine the impact of the number of variables on the performance of our method and other baselines. For more detailed information about the data used in each experiment, refer to Table~\ref{tab:safegraph-datasets} in the appendix.

\subsection{Baselines}
We compare the performance of BysGNN with three baseline groups: Naive Baselines, Static Graph Neural Networks, and Dynamic Graph Neural Networks. The Naive Baselines include two simple statistical models: Naive Seasonal and Historical Average. The Static GNNs group consists of ConvGRU~\cite{seo2018structured}, ConvLSTM~\cite{seo2018structured}, DCRNN~\cite{li2017diffusion}, and A3T-GCN~\cite{bai2021a3t} models, which operate on a predefined graph structure based on pairwise Euclidean distances between POIs. The Dynamic GNNs group includes StemGNN~\cite{cao2020spectral}, a state-of-the-art technique for MTS forecasting that uses an attention mechanism to infer relationships between nodes. Further details on each baseline model can be found in the appendix \ref{app:baselines}.

\begin{table*}[t]
\caption{Comparison of Forecasting Results for Different Datasets in Two Data Regimes: Large Data Regime (400 POIs) and Small Data Regime (40 POIs). The value on the left of the vertical bar at each field represents the error for the large data regime, while the value on the right represents the error for the small data regime. The lowest error value is highlighted in bold, while the second-lowest value is denoted in italics with underline. The ``Improvement'' row displays the percentage improvement in the relative error of BysGNN compared to the best-performing baseline.}
\label{tab:comp-results-main}
\resizebox{\textwidth}{!}{%
    \centering
    \begin{tabular}{|c|ccc|ccc|ccc|}
    \toprule
        \bf{Dataset} & ~ & Houston & ~ & ~ & Los Angeles & ~ & ~ & New York City & ~ \\ 
        \bf{Evaluation Metric} & MAE & MAPE & RMSE & MAE & MAPE & RMSE & MAE & MAPE & RMSE \\ 
        \hline
        Naïve Seasonal & 4.746 | 13.509 & \emph{\underline{0.664}} | \bf{0.534} & 18.166 | 48.304 & 2.934 | 7.346 & 0.752 | \bf{0.586} & 10.005 | 18.507 & 3.681 | 9.425 & 0.699 | \emph{\underline{0.540}} & 9.216 | 23.27 \\ 
        Historical Average & 8.860 | 35.323 & 0.783 | 0.766 & 26.911 | 77.0 & 4.388 | 15.838 & 0.786 | 0.756 & 11.729 | 28.589 & 6.555 | 25.208 & 0.770 | 0.744 & 22.018 | 65.272 \\ \hline
        ConvGRU & 6.415 | 17.640 & 2.539 | 0.920 & 20.179 | 38.728 & 3.781 | 8.474 & 3.139 | 1.311 & 10.905 | 17.386 & 4.605 | 11.200 & 1.851 | 1.102 & 17.028 | 28.041 \\
        ConvLSTM & 8.076 | 24.789 & 4.270 | 1.695 & 23.127 | 49.629 & 4.362 | 10.500 & 4.397 | 1.755 & 11.879 | 20.068 & 5.448 | 14.095 & 2.697 | 1.751 & 18.959 | 33.199 \\
        DCRNN & 5.683 | 16.371 & 1.990 | 0.778 & 18.941 | 36.803 & 3.389 | 7.870 & 2.693 | 1.232 & 9.879 | 16.433 & 4.139 | 10.395 & 1.605 | 0.940 & 15.504 | 26.247 \\ 
        A3T-GCN & 8.380 | 30.921 & 3.377 | 2.014 & 23.9 | 63.144 & 4.604 | 13.506 & 4.129 | 1.929 & 12.601 | 26.354 & 5.824 | 16.803 & 2.579 | 1.726 & 20.171 | 42.065 \\        
        \hline
        StemGNN & \emph{\underline{4.390}} | \emph{\underline{11.840}} & 0.735 | 0.786 & \emph{\underline{14.604}} | \emph{\underline{32.745}} & \emph{\underline{2.485}} | \emph{\underline{6.050}} & {\bf0.671} | 0.716 & \emph{\underline{6.951}} | \bf{10.552} & \emph{\underline{3.261}} | \emph{\underline{8.505}} & \emph{\underline{0.652}} | 0.683 & \emph{\underline{8.074}} | \emph{\underline{18.439}} \\ 
        \bf{BysGNN} & \bf{4.095} | \bf{11.095} & {\bf0.658} | \emph{\underline{0.648}} & \bf{12.904} | \bf{30.332} & \bf{2.377} | \bf{5.610} & \emph{\underline{0.676}} | \emph{\underline{0.608}} & {\bf6.091} | \emph{\underline{12.453}} & \bf{3.113} | \bf{6.863} & \bf{0.598} | \bf{0.478} & \bf{7.351} | \bf{15.450} \\
        \hline
         Improvement & +6.71\% | +6.29\% & +0.90\% | -21.34\% & +11.64\% | +7.36\% & +4.34\% | +7.27\% & -0.74\% | -3.75\% & +12.37\% | -18.01\% & +4.53\% | +19.30\% & +8.28\% | +11.48\% & +8.95\% | +16.21\% \\

        \bottomrule
        \end{tabular}
        }

        
\end{table*}

\subsection{Experiment Results}
We used input windows of 24 hours to train each GNN-based model to predict the number of visits for each POI for the next 6 hours for the datasets described in Table \ref{tab:safegraph-datasets} of the appendix. The forecasting results for each dataset for both large (400 POIs in each city) and small (40 POIs in each city) data regimes are presented in Table~\ref{tab:comp-results-main}. Although our experiments included visitor data from five different cities, Table \ref{tab:comp-results-main} presents results for only three cities due to space limitations. The results for the remaining two cities exhibit the exact same trend and are reported in Table~\ref{tab:comp-results-appendix} of the appendix.

The results of the large data regime presented in Table \ref{tab:comp-results-main} (values to the left of the vertical line for each field) show that our proposed BysGNN model consistently outperforms all other baselines across all datasets, except for the MAPE value on the Los Angeles dataset. This demonstrates the superiority of our architecture for forecasting seasonal time-series data. Moreover, our Dynamic GNN models (StemGNN and BysGNN) demonstrate significantly lower error values compared to Static GNN models. This underscores the limitation of using a static graph with predefined relationships between variables, even when incorporating a sophisticated GNN Block. Although StemGNN performs the best among the baselines, our BysGNN model outperforms it in almost every instance with an error reduction of up to 12.3\%, despite having a less complex GNN Block. This result validates our assumption that a well-designed Graph Construction Block would substantially enhance forecasting performance. Specifically, while StemGNN improved upon Static GNNs by constructing a dynamic graph solely based on the time-series windows, our BysGNN model took it a step further by introducing multi-context correlations that are more resilient to noise and yield a more precise depiction of the underlying relationships between variables. 

The values on the right of the vertical line for each field in Table~\ref{tab:comp-results-main} present the results for the small data regime, where there are only 40 POIs in each dataset. BysGNN continues to demonstrate the best overall performance, outperforming all other GNN-based models, including StemGNN, by up to 19\% error reduction in all cases except for RMSE for Los Angeles.
It is worth noting that all models exhibit worse RMSE and MAE results in this data regime compared to the large number of POIs data regime. This is because RMSE and MAE depend on the scale of the number of visits, and the average number of visits is significantly higher in the small data regime compared to the previous large data regime (see Table \ref{tab:safegraph-datasets}). In contrast, MAPE is not affected by the scale of the number of visits and provides a better measure to compare these two regimes.
As shown in Table \ref{tab:comp-results-main}, MAPE in static GNNs improves significantly in the small regime compared to the large data regime, indicating that strong predefined assumptions about the relationships between variables work better when the number of variables is smaller. StemGNN, on the other hand, is the only model that performs worse in every case in the small data regime compared to the large data regime. This highlights a major drawback of previous Dynamic GNNs: since they rely solely on the similarity of visit patterns to build a dynamic graph representing relationships, they require a high number of variables (nodes) to uncover meaningful relationships. Consequently, they may fail to infer accurate inter-node relationships based on the limited number of time series, such as in the small data regime.
We even observe that, in the case of Houston, a static GNN like DCRNN outperforms StemGNN in terms of MAPE. Conversely, BysGNN shows improved MAPE performance in all cases, underscoring the importance of accounting for multi-context correlations in graph construction.

Another interesting observation is the high effectiveness of the Naive Seasonal model, which outperforms most Static GNN models in both regimes. This is due to the highly seasonal nature of our visitation time-series datasets, where the weekly number of visits to most POIs remains relatively stable. Consequently, the Naive Seasonal model is a reasonable forecasting model and hard to beat. GNN-based models only have access to exact visiting numbers during the last 24 hours in the input sequence, making it more difficult for them to outperform the Naive Seasonal model. However, StemGNN and BysGNN are able to beat the Naive Seasonal forecasts due to the robustness of their dynamic graph, which accounts for the dynamic intra- and inter-time-series correlations at each window. This further highlights the significance of a flexible and expressive graph structure.

\begin{table*}[t!]
\caption{Ablation Study Results for the Houston Dataset. The percentage of relative error change for each variant compared to the original BysGNN is listed below the actual error value. The highest percentage of the error change for each evaluation metric is highlighted in bold.}
\label{tab:abl-study}
    \centering
\resizebox{0.8\textwidth}{!}{%
    \begin{tabular}{c||c|c|c|c|c|c}
    \toprule
        ~ & BysGNN & w/o. Semantics & w/o. Space & w/o. Meta-Nodes & w.o Self-Attention & w.o Adj-Thresholding\\
        \hline
        MAE & \textbf{3.916} & 4.561 & 4.209 & 4.341 & 4.283 & 4.348\\
         & -- & \textbf{+16.47}\% & +7.48\% & +10.85\% &  +9.37\% & +11.03\%\\
        \hline
        MAPE & \textbf{0.618} & 0.710 & 0.653 & 0.660 & 0.620 & 0.682\\
        & -- & \textbf{+14.88\%} & +5.66\% & +6.79\% & +0.32\% & +10.35\%\\
        \hline
        RMSE & \textbf{10.318} & 13.327 & 12.108 & 13.591 & 12.906 & 13.334\\
        & -- & +29.16\% & +17.34\% & \textbf{+31.72}\% & +25.08\% & +29.23\%\\

        \bottomrule
    \end{tabular}
    }
\end{table*}

\subsection{Ablation Study}~\label{ablation_study}

We created five variations of our proposed BysGNN model to understand the effectiveness of different BysGNN components: (1) \textbf{w.o Semantics}: BysGNN without utilizing POI semantics in node features and semantics similarities in adjacency matrix; (2) \textbf{w.o Space}: BysGNN without utilizing spatial correlations and distance between POIs in the Multi-Context Correlations Layer; (3) \textbf{w.o Meta-Nodes}: BysGNN without the Aggregated Data Generator module; (4) \textbf{w.o Self-Attention}: BysGNN without the proposed self-attention mechanism in the Intra-Series Correlation Layer; (5) \textbf{w.o Adj-Thresholding}: BysGNN without applying thresholding to the output of the gating mechanism for the construction of the adjacency matrix.

Table \ref{tab:abl-study} shows the evaluation results for each scenario. It clearly shows that all studied components contribute to an improvement in results, thus confirming our hypothesis that incorporating multi-context correlations to build a more expressive and robust dynamic graph has a significant impact on forecasting quality.
Interestingly, we observed that removing semantic embeddings has the largest overall impact on the three combined performance metrics. This aligns with intuition, as similar types of POIs, such as restaurants and coffee shops, tend to experience similar visit patterns in real-world data, and BysGNN can effectively capture and utilize this semantic similarity. Moreover, while considering spatial correlations improves forecasting accuracy, the impact of considering semantics is greater than that of spatial correlations. This is again intuitive, as we would expect similar types of POIs, such as restaurants, to have similar visits regardless of their location in a city (e.g., high visits during lunch hours), while the visitation patterns of close-by POIs might not necessarily be as correlated (e.g., a gas station with a nearby restaurant).

Although semantics have the highest impact on combined evaluation metrics, meta-nodes seem to have the largest impact on RMSE alone. This indicates that capturing the multi-context dependencies between POI nodes and meta-nodes (taxonomic correlations) can significantly enhance the results. This is because, in POI visits datasets, individual POIs often follow a higher-level aggregated visit pattern. For instance, a single school would likely follow the visitation pattern of all schools at a higher level. Hence, meta-node patterns guide more precise learning of POI node patterns, and BysGNN can dynamically learn these relationships during training. Moreover, our hypothesis that learned attention scores based on visit patterns might be noisy and require the gated attention mechanism and a robust thresholding mechanism to diminish the impact of noisy patterns is supported by the substantial impact of adjacency thresholding on performance. Additionally, while the self-attention unit in the Intra-Series Correlation Layer improves the results, it has the least impact on MAPE values. This can be attributed to the fact that GRUs already exhibit a strong capability to capture intra-series dependencies.

\subsection{Interpretation of Adjacency Matrix}~\label{adj_interpret}
This section compares BysGNN's dynamically generated graphs with 1) static graphs defined using spatial and semantics similarity and 2) dynamic graphs using attention on time-series windows for their forecasting performance (Section \ref{sec:adj-perf-impact}) and impact on node embeddings (Section \ref{sec:adj-node-embeds}). For dynamic graphs, we use the graphs created on Wednesday, Jan 22, 2020, at 10:00:00 AM, representing a typical weekday scenario. To simplify visualization, we show the results from the top $40$ most visited POIs in NYC from $9$ POI categories, resulting in $10$ meta-nodes (one for each of the $9$ categories and $1$ for all POIs, the $global$ meta-node).


\vspace{-5pt}

\begin{table}[htbp]
  \centering
  \caption{Error Results for Different Adjacency Matrices}
  \vspace{-5pt}
  \label{tab:diff-adj-res}
  \resizebox{0.5\textwidth}{!}{%
  \begin{tabular}{lcccc}
    \toprule
    & \textbf{BysGNN's Original} & Temporal Attention & Spatial Similarities & Semantics Similarity \\
    & \textbf{Adjacency Matrix} & Matrix & Matrix & Matrix \\
    \midrule
    MAPE & \textbf{0.455} & \emph{\underline{0.505}} & 0.562 & 0.560 \\
    MAE & \textbf{6.673} & \emph{\underline{7.033}} & 8.050 & 7.830
    \\
    RMSE & \textbf{13.962} & \emph{\underline{15.677}} & 19.436 & 17.463
    \\
    \bottomrule
  \end{tabular}
  }
\end{table}

\vspace{-10pt}

\subsubsection{Impact of Graph Types on Forecasting Performance}
\label{sec:adj-perf-impact}
Table \ref{tab:diff-adj-res} shows that BysGNN's dynamically generated graphs from multiple contexts consistently outperformed the other graphs, indicating its effectiveness in capturing relevant temporal relationships while preserving static relationships among nodes. The dynamic graphs derived from temporal attention yielded the second-best performance, highlighting the effectiveness of dynamic GNNs in forecasting. Using spatial similarity performed worse than semantics similarity, especially in RMSE, suggesting that POI proximity alone is not a good indicator for capturing meaningful visiting pattern relationships. The following section will examine specific examples to showcase the node embeddings learned from these graphs.

\begin{figure*}[t!]
    \centering
    \subfloat[Spatial Similarities Matrix]{%
        \includegraphics[width=0.25\textwidth]{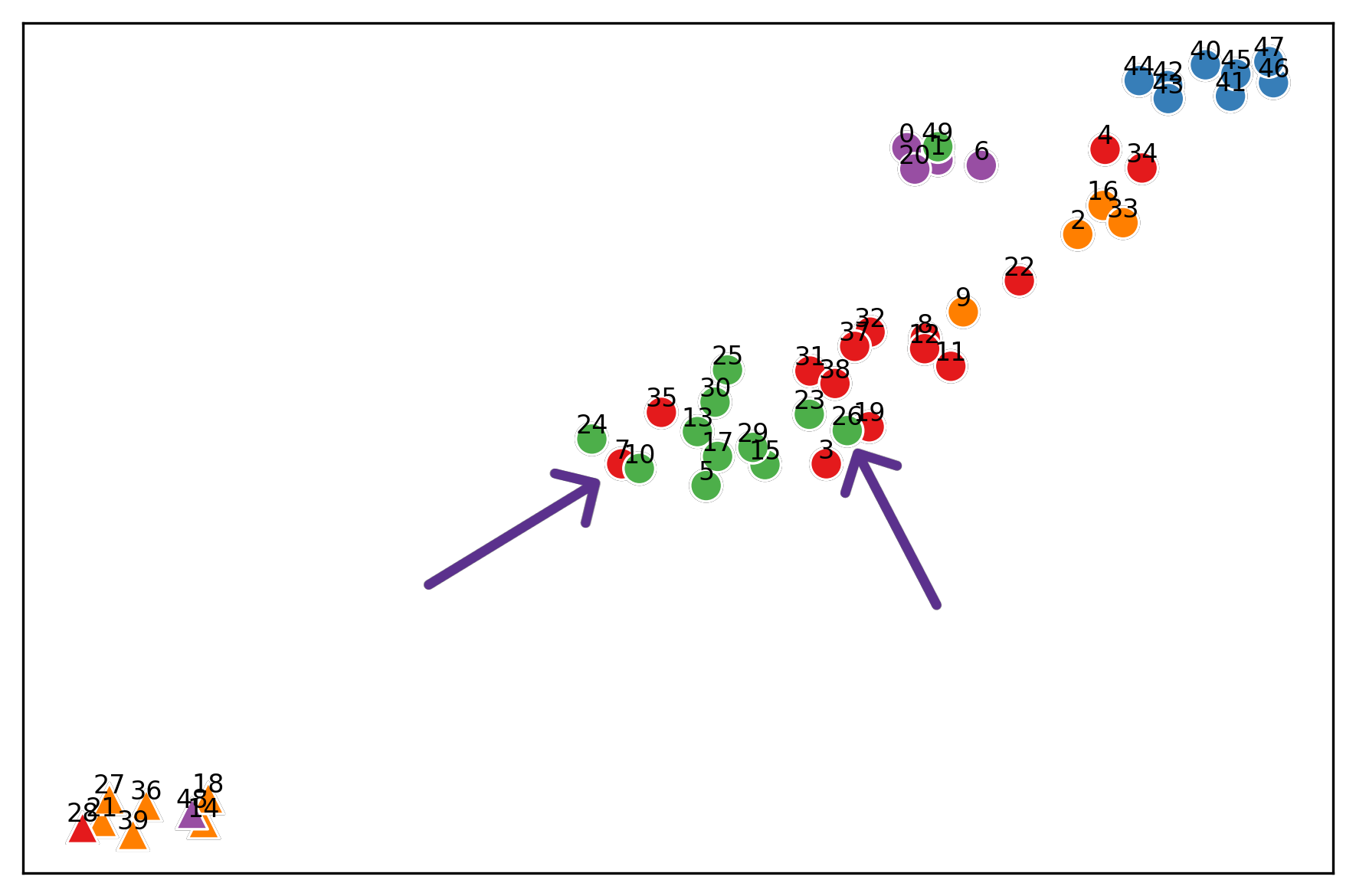}%
        \label{fig:embedding_spatial_mat}%
    }\hfill
    \subfloat[Semantics Similarity Matrix]{%
        \includegraphics[width=0.25\textwidth]{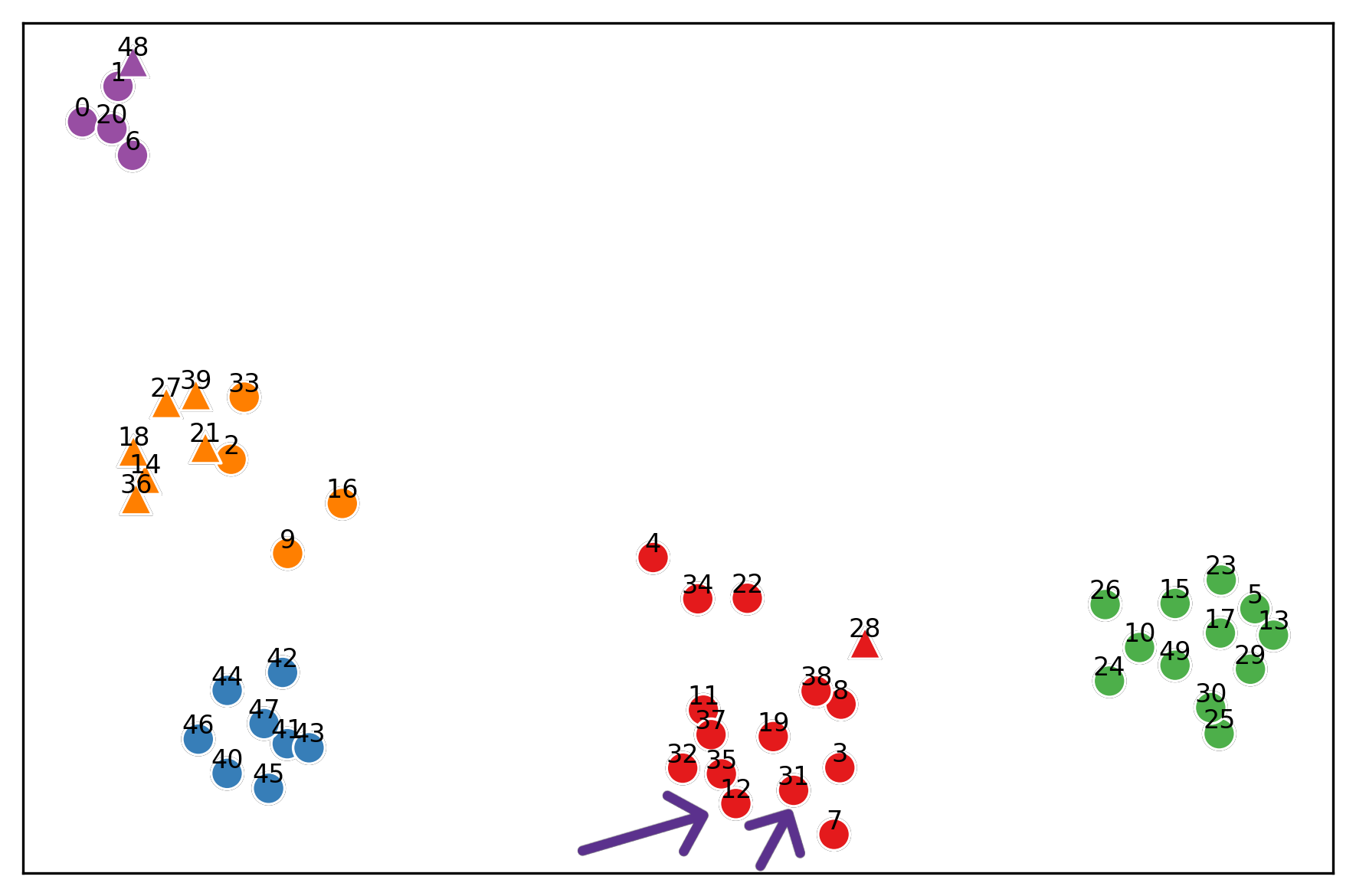}%
        \label{fig:embedding_semantics_mat}%
    }\hfill
    \subfloat[Temporal Attention Matrix]{%
        \includegraphics[width=0.25\textwidth]{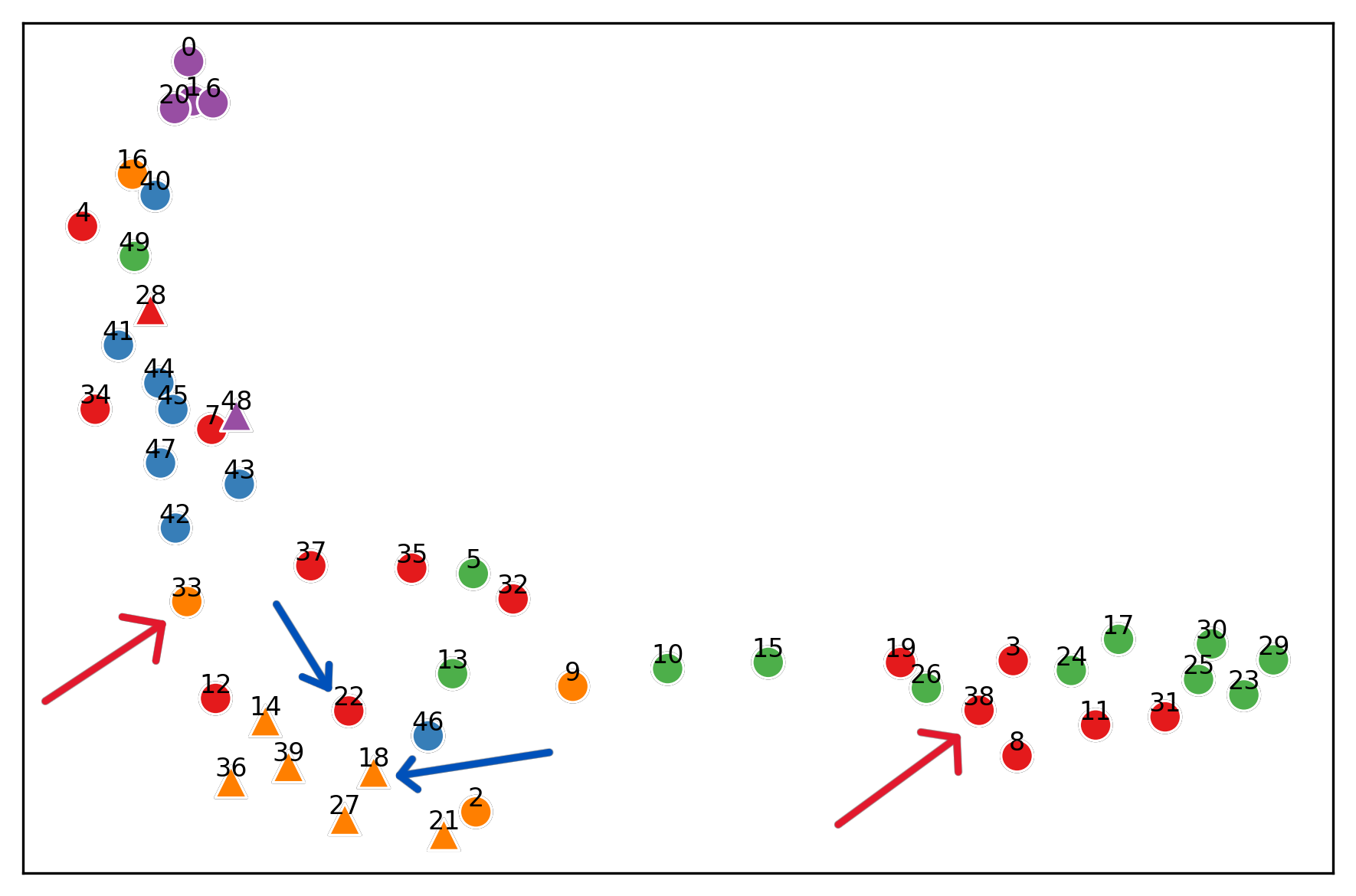}%
        \label{fig:embedding_attention_mat}%
    }\hfill
    \subfloat[Busyness Graph's Fused Adjacency Matrix]{%
        \includegraphics[width=0.25\textwidth]{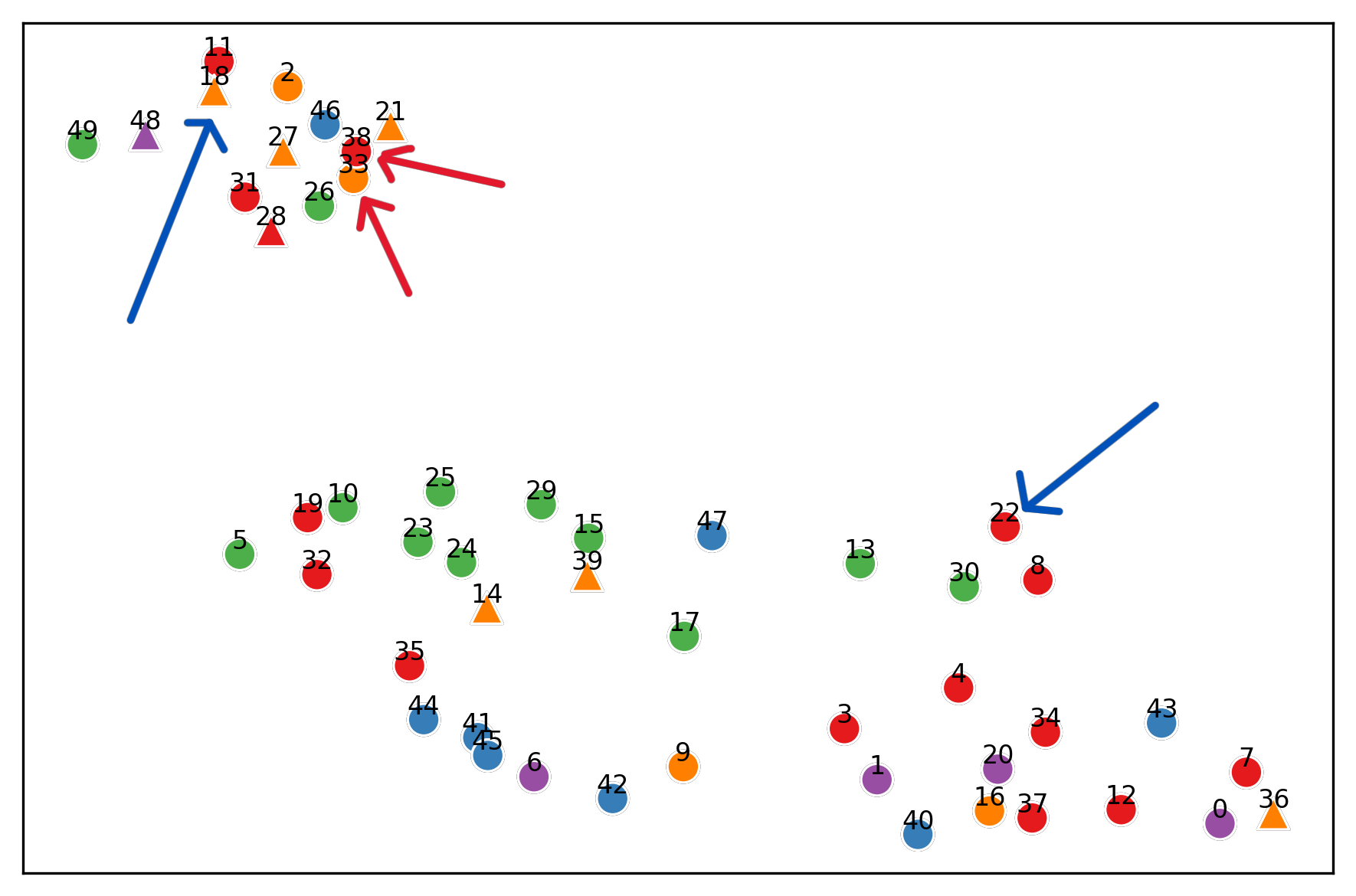}%
        \label{fig:embedding_bysgnn_adj}%
    }
    \vspace{-10pt}
    \caption{Visualization of Node Embeddings for Different Graphs Obtained After the GNN Block. The embeddings are projected to the 2D space using the UMAP dimension reduction technique. Each data point represents a node or meta-node, with semantic embedding clusters depicted by five distinct colors and spatial embedding clusters indicated by two different shapes (triangles and circles). The numbers atop the points correspond to the node indices.}
    \label{fig:embeddings}
\end{figure*}

\begin{figure*}[tbp]
\vspace{-10pt}
    \centering
    \subfloat[POIs \emph{33} and \emph{38} exhibit a high similarity in time series patterns beyond the most recent observed window (yellow). These POIs are situated close to each other in Busyness Graph's embedding space, while they are far apart when using the temporal attention matrix.]{%
        \includegraphics[width=0.49\textwidth]{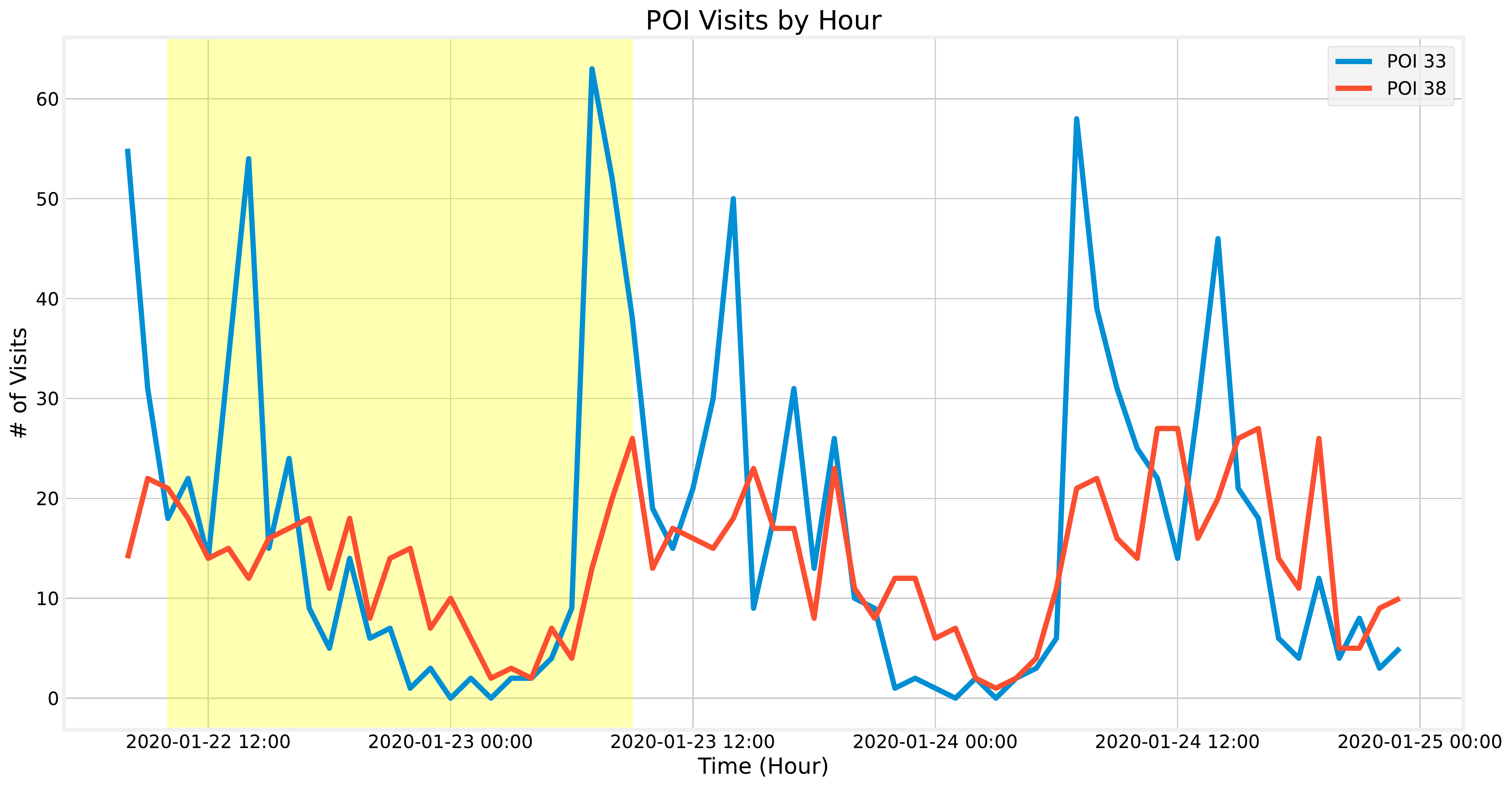}%
        \label{fig:pois33-38-ts}%
    }\hfill
    \subfloat[POIs \emph{22} and \emph{18} show dissimilar future visit given the highlighted input window (yellow). These POIs are situated far from each other in the embedding space using Busyness Graph's  adjacency matrix, while they are close using the temporal attention matrix.]{%
        \includegraphics[width=0.49\textwidth]{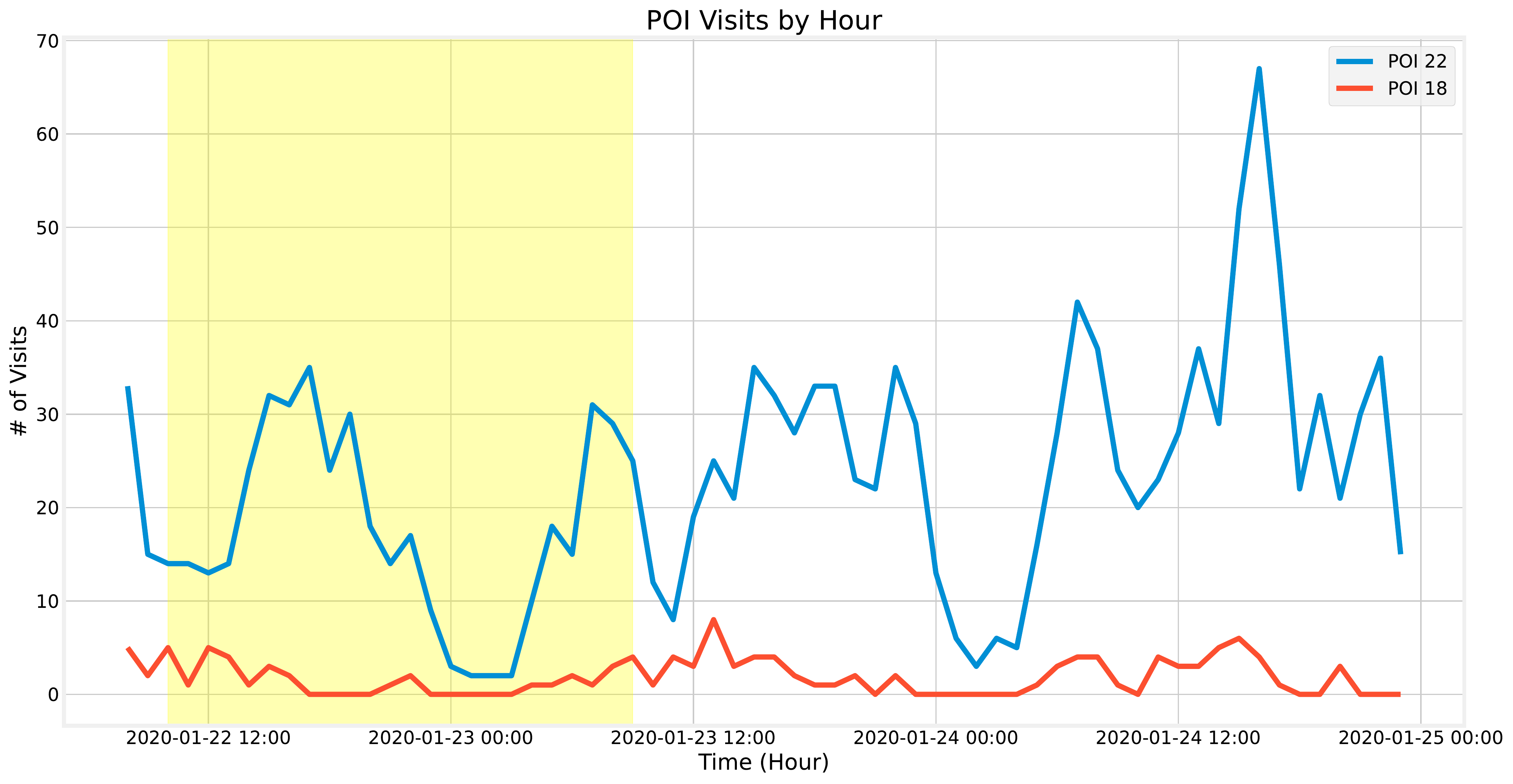}%
        \label{fig:pois22-18-ts}%
    }\hfill
    \vspace{-10pt}
    \caption{Visits Time Series for Two Pairs of POIs Within and Beyond the Observed Input Window (Highlighted in Yellow)}
    \label{fig:res-series}
\end{figure*}


\subsubsection{Impact of Graph Types on Node Embeddings}
\label{sec:adj-node-embeds}
Figure \ref{fig:embeddings} illustrates the node embeddings obtained after the GNN Block, projected onto a 2D space using UMAP~\cite{mcinnes2018umap} dimensionality reduction technique for each graph type in Table~\ref{tab:diff-adj-res}. Each data point represents a node or a meta-node, with two shapes (triangles and dots) representing the spatial embedding clusters (Figure \ref{fig:embedding_spatial_mat}) and five colors indicating the semantic embedding clusters (Figure \ref{fig:embedding_semantics_mat}).

The two clusters of spatial embeddings correspond to two boroughs of the 40 POIs: Manhattan and Queens. For instance, POI 7 and 26 in Figure \ref{fig:embedding_spatial_mat}, highlighted with purple arrows, represent ``Red Mango'' (yogurt shop) and ``42nd Street'' (iconic crosstown street), respectively. Despite being in Manhattan within a half-mile distance, their visit patterns differ significantly, demonstrating geospatial proximity does not always indicate similar visit patterns between POIs. 

The five clusters of semantic embeddings exhibit oversmoothing~\cite{chen2020measuring}, discard minor variations in visiting patterns between POIs, and potentially lead to similar forecasting results for POIs in the same cluster. Consider POIs 12, "Theodore Roosevelt Park", and 31, "Herald Square", indicated by the purple arrows in Figure \ref{fig:embedding_semantics_mat}. They are in the same semantic cluster due to their shared POI category of "historical sites." However, they have substantially different visit trends, and relying solely on semantic similarities also fails to capture the relationship between their visit patterns.

Figures \ref{fig:embedding_attention_mat} and \ref{fig:embedding_bysgnn_adj} illustrate the node embeddings from the temporal attention and BysGNN graphs. In both embedding spaces, neighboring embeddings do not have consistent spatial and semantic similarity (neighbors have different colors and shapes). Also, BysGNN's embeddings result in two clusters, while the node embeddings from the temporal attention graph do not exhibit clear cluster boundaries. 

Consider nodes 33 and 38 (highlighted with red arrows), which are nearby in the BysGNN's embedding space while far away in the embedding space learned from the temporal attention graph. The recent visits window to these POIs is highlighted in yellow in Figure \ref{fig:pois33-38-ts}, and the model aims to forecast the visits to the right of this highlighted area. Although the blue and red sequences in the highlighted window share a similar trend, the temporal attention graph assigns distant embeddings for these two nodes. This happens because the temporal attention mechanism is optimized for the entire sequence rather than for the subsequence in the highlighted window. In contrast, BysGNN effectively captures this similarity by considering additional contextual information. As this pair of POIs belongs to the same spatial cluster, BysGNN enhances the similarity score derived from temporal attention for this specific pair, resulting in close node embeddings. 

On the other hand, let us focus on nodes 22 and 18, highlighted with blue arrows in Figures \ref{fig:embedding_attention_mat} and \ref{fig:embedding_bysgnn_adj}. Here, the temporal attention graph would predict a similar visit pattern for the input visit windows depicted in Figure \ref{fig:pois22-18-ts}, due to the same aforementioned reasons. However, BysGNN considers additional contexts: these POIs belong to different spatial and semantic clusters. Consequently, BysGNN accurately determines that their future visit patterns should be dissimilar and learns distant node embeddings for them.


These findings underscore the shortcomings of relying exclusively on predefined spatial and semantic relationships or only on dynamic time-series windows to capture accurate visit pattern correlations. In contrast, BysGNN provides a robust solution by effectively considering all contexts.

\section{Conclusion}\label{conclusion}
This work presents BysGNN, a dynamic graph neural network that is specifically designed to uncover multi-context correlations among POIs for accurate visit forecasting. Using various sources of information, including geographic information, visit numbers, semantics, and taxonomic information of POIs, BysGNN learns an accurate dynamic graph representation that is then passed to a simple GNN block for forecasting. Our experiments on real-world datasets across the United States demonstrate the superiority of BysGNN over state-of-the-art forecasting models, including those using highly sophisticated GNN blocks. In future work, we plan to apply BysGNN to other datasets with similar underlying multi-context correlations, such as health data.


\begin{acks}
Research supported by the National Science Foundation (NSF) under CNS-2125530, the National Institute of Health (NIH) under grant 5R01LM014026, the Intelligence Advanced Research Projects Activity (IARPA) via the Department of Interior/Interior Business Center (DOI/IBC) contract number 140D0423C0033 and NURI contract HM04762210001 (approved for public release under NGA-U-2023-00192). The U.S. Government is authorized to reproduce and distribute reprints for Governmental purposes, notwithstanding any copyright annotation thereon. Disclaimer: The views and conclusions contained herein are those of the authors and should not be interpreted as necessarily representing the official policies or endorsements, either expressed or implied, of IARPA, DOI/IBC, NSF, NIH, NGA or the U.S. Government.
\end{acks}

\bibliographystyle{ACM-Reference-Format}
\bibliography{references}

\clearpage

\appendix
\section{Appendix}



\begin{table*}[tbp]
\caption{Summary of Datasets used in Experiments}
\label{tab:safegraph-datasets}
\resizebox{\textwidth}{!}{
\begin{tabular}{|c|c|c|c|c|c|c|}
\toprule
\textbf{City} & \multicolumn{2}{c|}{\textbf{Number of POIs}} & \multicolumn{2}{c|}{\textbf{Number of POI Categories}} & \multicolumn{2}{c|}{\textbf{Average Number of Hourly Visits}} \\
\cline{2-7}
& \textbf{Large Regime} & \textbf{Small Regime} & \textbf{Large Regime} & \textbf{Small Regime} & \textbf{Large Regime} & \textbf{Small Regime} \\
\hline
Houston & 400 & 40 & 43 & 12 & 11 & 48 \\ \hline
Chicago & 400 & 40 & 35 & 14 & 6 & 22 \\ \hline
Los Angeles & 400 & 40 & 44 & 14 & 6 & 23 \\ \hline
New York City & 400 & 40 & 32 & 10 & 9 & 36 \\ \hline
San Antonio & 400 & 40 & 42 & 12 & 8 & 28 \\
\bottomrule
\end{tabular}
}
\end{table*}

\begin{table*}[]
\caption{Comparison of Forecasting Results for Chicago and San Antonio Datasets in Two Data Regimes: Large Data Regime (400 POIs) and Small Data Regime (40 POIs). The value on the left of the vertical bar at each field represents the error for the large data regime, while the value on the right represents the error for the small data regime. The lowest error value is highlighted in bold, while the second-lowest value is denoted in italics with underline. The ``Improvement'' row displays the percentage improvement in the relative error of BysGNN compared to the best-performing baseline.}
\label{tab:comp-results-appendix}
    \centering
    \resizebox{\textwidth}{!}{%
    \begin{tabular}{|c|ccc|ccc|}
    \toprule
        \bf{Dataset} & ~ & Chicago & ~ & ~ & San Antonio & ~ \\
        \bf{Evaluation Metric} & MAE & MAPE & RMSE & MAE & MAPE & RMSE \\ \hline
        Naïve Seasonal & 3.237 | 8.502 & 0.754 | \bf{0.661} & \emph{\underline{9.216}} | 31.17 & 3.85 | 9.595 & 0.689 | \emph{\underline{0.564}} & 10.573 | 21.331 \\
        Historical Average & 4.624 | 15.428 & 0.791 | 0.782 & 14.011 | 35.517 & 6.494 | 21.845 & 0.78 | 0.776 & 15.776 | 39.095 \\ \hline
        ConvGRU & 4.18 | 11.558 & 2.675 | 1.319 & 10.753 | 25.919 & 4.622 | 10.491 & 2.838 | 0.853 & 10.218 | 19.909 \\
        ConvLSTM & 5.019 | 14.811 & 3.853 | 1.927 & 11.972 | 30.275 & 5.745 | 13.573 & 5.202 | 1.617 & 11.608 | 23.695 \\
        DCRNN & 3.756 | 10.808 & 2.327 | 1.149 & 9.857 | 24.235 & 4.167 | 10.137 & 2.269 | 0.827 & 9.513 | 19.091 \\
        A3TGCN & 5.343 | 18.698 & 3.654 | 2.268 & 12.821 | 38.289 & 6.086 | 18.594 & 4.517 | 2.357 & 12.731 | 30.877 \\
        \hline
        StemGNN & \emph{\underline{2.776}} | \emph{\underline{7.891}} & \emph{\underline{0.724}} | 1.014 & 9.857 | \emph{\underline{18.930}} & \emph{\underline{3.371}} | \emph{\underline{9.432}} & \emph{\underline{0.686}} | 0.757 & \emph{\underline{8.923}} | \emph{\underline{17.994}} \\
        \bf{BysGNN} & {\bf2.750} | \bf{6.541} & {\bf0.718} | \emph{\underline{0.733}} & {\bf8.218} | \bf{15.867} & {\bf3.278} | \bf{7.759} & {\bf0.629} | \bf{0.533} & {\bf8.418} | \bf{16.352}\\
        \hline
        Improvement & +0.93\% | +17.10\% & +0.82\% | -10.89\% & +10.82\% | +16.18\% & +2.75\% | +17.73\% & +8.30\% | +5.49\% & +5.65\% | +9.12\%\\
        
\bottomrule
    \end{tabular}
    }
\end{table*}

\subsection{Experiment Details}
\label{app:exp-details}
\subsubsection{Baselines}
\label{app:baselines}
We compare the performance of BysGNN with three groups of baselines:
\begin{itemize}
    \item \textbf{Naive Baselines}: We first compare our model with two simple statistical baselines that provide relatively accurate results when the visitation time series is highly seasonal. (1) \textbf{Naive Seasonal}: We use the number of visits on the same day/time from the previous week as the prediction values for the same day/time of the current week. (2) \textbf{Historical Average}: We take the average of the number of visits for the same day of the week during the previous month as the prediction, similar to Google Maps' popular times graph~\cite{google-busyness}.
    \item \textbf{Static Graph Neural Networks}: These GNN-based models operate on a predefined graph structure and, therefore, require prior knowledge of the graph topology. For our experiments, we construct a static graph based on the pairwise Euclidean distance between different POIs. (1) \textbf{ConvGRU} and (2) \textbf{ConvLSTM} \cite{seo2018structured}: these models combine a GRU and an LSTM, respectively, with ChebNet \cite{defferrard2016convolutional} to make spatiotemporal forecasts. (3) \textbf{DCRNN} \cite{li2017diffusion}: This model adopts an encoder-decoder framework and proposes a diffusion convolutional layer to capture the spectral and temporal dependencies. (4) \textbf{A3T-GCN} \cite{bai2021a3t}: This model captures the global temporal dynamics and spatial correlations given a graph structure. Moreover, it introduces an attention mechanism to adjust the importance of different time points and boost forecasting accuracy. 
    \item \textbf{Dynamic Graph Neural Networks}: This family of GNN-based models does not require a predefined graph structure but instead uses an attention mechanism to infer relationships between nodes. (1) \textbf{StemGNN}: This method~\cite{cao2020spectral} combines the Graph Fourier Transform and Discrete Fourier Transform to learn the correlations among the time series of different nodes. This is a state-of-the-art (SOTA) technique for MTS forecasting.
\end{itemize}


\subsubsection{Hardware and Software Setup}

Our experiments were performed on a cluster node equipped with an 18-core Intel i9-9980XE CPU, 125 GB of memory, and two 11 GB NVIDIA GeForce RTX 2080 Ti GPUs. Furthermore, all neural network models are implemented based on PyTorch version 1.13.0 with CUDA 11.7 using  Python version 3.10.8. We also implemented the GNN-based baselines (with the exception of StemGNN) using the Pytorch Geometric Temporal library~\cite{rozemberczki2021pytorch}.

\subsubsection{Evaluation Metrics}
Since we modeled the problem of forecasting the number of visits to POIs as a time-series forecasting task, we evaluate our prediction performance by comparing the average of Mean Absolute Error (MAE), Mean Absolute Percentage Error (MAPE), and Root Mean Squared Error (RMSE) over the prediction horizon ($H$ timesteps). The smaller values for these metrics indicate better forecasting performance. 
Here is the description of each metric:
\begin{itemize}
	\item \textbf{MAE:} Average of the difference between the ground truth and the predicted values.
	\item \textbf{MAPE:} The percentage equivalent of MAE.
	\item \textbf{RMSE:} The square root of the average of the squared difference between the ground truth and the predicted values.
\end{itemize}


\subsubsection{Hyperparameter Configuration}
The hyperparameters used in our model were carefully chosen using cross-validation to optimize performance. Here is a summary of the key configurations:
\begin{itemize}
    \item \textbf{Dataset Split}: We divided the dataset into three parts, allocating $70\%$ for training, $20\%$ for validation, and $10\%$ for testing. The respective time periods for each set were $280$ days, $80$ days, and $40$ days.
    \item \textbf{Data Normalization}: Z-score normalization was applied to ensure standardized input data.
    \item \textbf{Training}: The model was trained using the RMSProp optimizer with an initial learning rate of $0.001$. The learning rate decayed by a factor of $0.2$ every $10$ epochs. We trained the model for $40$ epochs with a batch size of $32$.
    \item \textbf{Adjacency Thresholding}: A case amplification factor of $2.5$ was employed to enhance the performance of the adjacency thresholding module. The threshold value $\eta$ was set to $0.15$, determining the cutoff for adjacency values.
    \item \textbf{Embedding Dimensions}: The temporal embedding dimension $M$ was set to $128$, and the semantics embedding dimension $P$ was set to $168$.
    \item \textbf{Attention Heads}: The Inter-Series Correlation Layer used a Multi-Head Attention mechanism with $8$ heads.
    \item \textbf{GNN Node Embedding}: The node embedding dimension for graph convolution in the GNN block, denoted as $M^*$, was set to $32$.
    \item \textbf{Gaussian Kernel Threshold}: The threshold value $\tau$ for the Gaussian kernel was determined as twice the standard deviation of distances between POIs, varying depending on the specific dataset used in each experiment.
\end{itemize}
All hyperparameters were kept consistent across all experiments except for the Gaussian kernel threshold.



\end{document}